\documentclass{article}
\usepackage{spconf,amsmath,graphicx}

\title{Three-Dimensional Segmentation of Vesicular Networks of Fungal Hyphae in Macroscopic Microscopy Image Stacks}
\name{P. Saponaro, W. Treible, A. Kolagunda, S. Rhein, J. Caplan, C. Kambhamettu, R. Wisser }
\address{University of Delaware, Newark, DE}
\begin{document}
%
\maketitle
\begin{abstract}
Automating the extraction and quantification of features from three-dimensional (3-D) image stacks is a critical task for advancing computer vision research. The union of 3-D image acquisition and analysis enables the quantification of biological resistance of a plant tissue to fungal infection through the analysis of attributes such as fungal penetration depth, fungal mass, and branching of the fungal network of connected cells. From an image processing perspective, these tasks reduce to segmentation of vessel-like structures and the extraction of features from their skeletonization. In order to sample multiple infection events for analysis, we have developed an approach we refer to as macroscopic microscopy. However, macroscopic microscopy produces high-resolution image stacks that pose challenges to routine approaches and are difficult for a human to annotate to obtain ground truth data.  We present a synthetic hyphal network generator, a comparison of several vessel segmentation methods, and a minimum spanning tree method for connecting small gaps resulting from imperfections in imaging or incomplete skeletonization of hyphal networks. Qualitative results are shown for real microscopic data. We believe the comparison of vessel detectors on macroscopic microscopy data, the synthetic vessel generator, and the gap closing technique are beneficial to the image processing community.

\end{abstract}
\vspace{-.5em}
\section{Introduction}
\vspace{-1em}
\label{sec:intro}
Segmentation of vessel-like structures are important for many applications including analysis of fungal infection of plant tissue \cite{minker2016semiautomated}, retinal vessel segmentation for systemic disease detection \cite{6724460}, and apple tree branch segmentation for a harvesting robot \cite{Ji201611173}. The focus of this study is on the automated segmentation of fungal infection networks (connected strands of cells that form branched networks) in macroscopic microscopy images of maize. Macroscopic microscopy is a term used to refer to microscopy images that are stitched together to produce an image of a macroscopic specimen in microscopic detail \cite{minker2016semiautomated}. The high resolution of microscopy image stacks, combined with artifacts from fluorescent staining and image processing pose a challenge to obtaining accurate segmentations in an efficient manner.

Moreover, this data is cumbersome to segment for a human, such that no ground truth data exist to quantitatively compare different algorithms. It is difficult for a human to manually segment for a few reasons: existing tools are not well suited for 3-D segmentation (requiring slice-by-slice editing), high intensity regions can be blurred between multiple Z-slices, non-uniform staining can cause strands of fungus to vary in intensity, and background artifacts from plant material can obscure the fungus. In order to compare algorithms for the segmentation of hyphal networks, we developed a synthetic image stack generator that simulates vessel-like networks.

Even with precise segmentations, small gaps can be present due to nonuniform staining artifacts in the underlying image data. This can create more hyphal networks than actually exist, leading to incorrect estimates of the length, depth, and branching features of true networks. To solve this problem, we employ a minimum spanning tree (MST) algorithm to backfill small gaps in a hyphal network. We implement this algorithm in a generalized way to work with N-dimensional (N-D) data.

The contribution of this work is as follows.
\begin{itemize}
\item Development of an open source synthetic hyphal network generator to quantitatively compare segmentation algorithms.
\item Development of a skeleton gap-filling method that works for N-dimensional data.
\item Quantitative and qualitative comparison of many vessel segmentation schemes on both synthetic and real macroscopic microscopy data.
\end{itemize}
%
%
%

The rest of this paper is organized as follows. Section \ref{sec:background} gives an overview of vessel segmentation approaches. Section \ref{sec:skel_gap} details the method for closing gaps in the skeletonized segmentation via MST. Section \ref{sec:exp_res} describes our experiments, Section \ref{sec:syn} presents the synthetic hyphal network generator along with quantitative results on synthetically generated data. Section \ref{sec:real} is a description and qualitative comparison of real macroscopic microscopy data from \cite{minker2016semiautomated}. Finally, the paper is summarized and concluded in Section \ref{sec:con}.

\section{Background}
\label{sec:background}
\vspace{-1em}
The segmentation of vesicular structures  including hyphal networks is a well-studied problem in which a variety of image processing approaches have been developed, such as active contours, edge approaches, filters for enhancing structures, and watershed approaches to name a few. For instance, Wildenhain et al. \cite{mqw} performed a Canny algorithm combined with morphological operations followed by removal of structures according to rules about the size, intensity, and shape of the object. Inglis and Gray \cite{CIS-171250} compared semi-automatic approaches and concluded that active contours performed best on fungal hyphae. Cai et al. \cite{doi:10.1117/12.834087} and Obara et al. \cite{doi:10.1093/bioinformatics/bts364} used a watershed transformation approach for segmentation.

For more general vessel segmentation, region-based approaches have been tried \cite{7160191}. Approaches have included texture-based descriptors such as local binary patterns (LBP) \cite{7796176}, color-based transformation and adaptive thresholding \cite{Ji201611173}, and an ant colonization optimization scheme \cite{7800141}. More recently, deep learning was used for retinal vessel detection \cite{7532384, DBLP:journals/corr/MajiSMS16}. This provides an overview about the wide variety of image processing approaches developed for segmentation, but this is by no means a comprehensive list of all publications on the subject. At least some of the approaches used for vessel segmentation are expected to be applicable to our macroscopic microscopy data of hyphal infection networks. In this study, we compare the effectiveness of different methods on our image data.

\section{Connecting The Skeleton Via Minimum Spanning Tree}
\label{sec:skel_gap}
\vspace{-1em}
The input to this module is a binary image stack containing the 3-D skeletonized segmentation of separate hyphal networks with small gaps due to segmentation errors or low image contrast. The output is a 3-D skeletonized segmentation with the gaps connected. We accomplish this using a minimum spanning tree (MST) algorithm \cite{Kruskal1956}. Congruent with the growth properties of hyphal networks, the MST algorithm does not create loops in the output skeleton. 

We formulate and implement this algorithm for an N-D binary array $S$. Define a "noxel" as an N-D point or N-D voxel. A graph $G$ representation of $S$ is constructed where the nodes are noxels of the skeleton, and the edges connect pairs of nodes. The weights of the edges are given by a distance function between the nodes, which is scaled in each direction by a scale factor $s_i$ (default is $s_i=1$). For noxels that reside in the same subgraph, the edge weight is set to $0 < \epsilon << 1$. For pairs of non-endpoints in separate subgraphs, the edges are severed. 
%
%
%
%

%

The MST algorithm \cite{Kruskal1956} is performed on graph $G$, which has the effect of connecting endpoints of a subgraph to their closest point in a neighboring subgraph. To restrict these new edges we introduce a gap length parameter $l$, and only accept edges of length $< l$. The final part of the algorithm is to rasterize the new edges back onto the image stack $S$. We accomplish that using Bresenham's algorithm \cite{Bresenham:1965:ACC:1663347.1663349}. 

\subsection{Implementation Details}
\vspace{-.5em}
Our implementation of this module includes an N-D endpoint detector, an N-D array to graph converter, and an N-D version of Bresenham's algorithm. All parts were implemented in vectorized Matlab code \footnote{https://github.com/drmaize/compvision/tree/master/xxxxx [to be released following publication]}. 

To detect N-D endpoints, we create a length 3 hypercube filter with 1 everywhere except the center, which is set to $3^N+1$. This filter is convolved with the N-D binary array $S$ to create $S^{\prime}$. Endpoints are defined at $S^{\prime} = 3^N+2$. The reason to set the center to $3^N+1$ is to exclude possible convolutions that reach the value of the center noxel without actually having the center noxel activated.

The distance function is defined as the geodesic time algorithm in \cite{Soille19941235}. That is, for a straight line path between $n_i, n_j$ containing noxels $n_k$,  
\vspace{-1em}
\begin{equation}
\vspace{-1em}
D(n_i, n_j) = 1-(\frac{S(n_i)}{2} + \frac{S(n_j)}{2} + \sum_{k=i+1}^{k=j-1} S(n_k)).
\end{equation}
 Alternative distance functions such as Euclidean distance may also be used.

To implement Bresenham's algorithm in N-D, the direction of the line between noxels $P$ and $Q$ is given by $P_i - Q_i$, the number of steps is given by $max(P_i - Q_i)$, and the step size is given by $\frac{P_i - Q_i}{max(P_i - Q_i)}$. 

Due to the vectorized implementation, the skeleton connection algorithm typically runs in about 8s on 2315x2378x150 3-D image stacks on our Sager Laptop containing an Intel Core i7-6400k CPU and 32GB of RAM. This can be further improved by breaking up large image stacks into smaller chunks and processing each in parallel. 

%
%

%

\begin{figure*}[!htbp]
\centering
\includegraphics[width=0.77\textwidth]{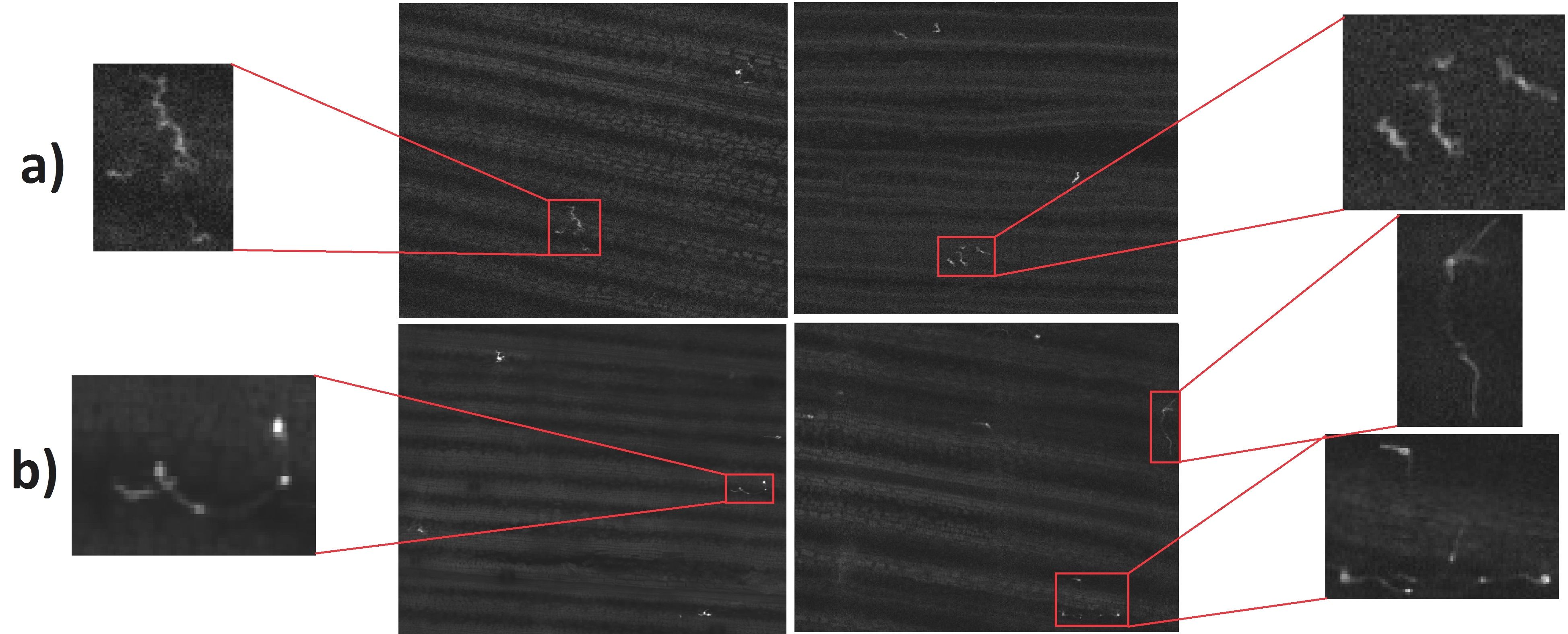}
\caption{Comparison of image slices from synthetically generated hyphal networks and real data. \textbf{a)} The top row contains synthetically generated image stacks. \textbf{b)} The bottom row contains real data.}
\label{syn_vs_real}
\end{figure*}

\section{Experiments and Results}
\label{sec:exp_res}
\vspace{-1em}
Due to the large and complex 3-D structure of hyphal networks captured by macroscopic microscopy, ground truth segmentation is hard to create from the data.  In order to circumvent this issue, we created a synthetic 3-D hyphal network generator that produces image stacks resembling real data. 

\vspace{-1em}
\subsection{Quantitative Analysis with Synthetically Generated Images}
\label{sec:syn}
\vspace{-1em}
Since the following section contains many random numbers and tunable parameters, we will release our code to ensure reproducibility \footnotemark[1]. 

\subsubsection{Synthetic Fungus Generator}
To generate a synthetic hyphal network, we implemented a weighted random walk algorithm. A synthetic hyphal network is created as a tree with each node having a probability of changing direction or branching in a direction. Branching and direction changing probabilities were chosen that approximate real hyphal networks from manual inspection. Specifically, there is a higher chance of in-plane fungal movement, a medium probability of splitting or moving down, and a very low probability of moving upward in the tissue. When a directional change is chosen, the new direction is selected such that the fungus will not "curl back" on itself. This is done by taking the dot product of the previous vector of the fungal path and the new candidate direction and ensuring that it is greater than zero.

Once the 3-D hyphal network locations are generated, the task is to transform that into an image stack. For each generated branch, we simulate non-uniform staining by varying the intensity according to a Piecewise Cubic Hermite Interpolating Polynomial (PCHIP) \cite{kahaner_moler_nash_1988} with randomly generated control points varying the intensity between [0.5,1]. Next, to simulate a defocus effect that causes the simulated hyphae to be blurred between Z-slices, we apply a 3D Gaussian filter with $\sigma=0.75$. To add artifacts from plant material that obscures the fungus, we take real microscopy images from our macroscopic microscopy data. Since the majority of the data lacks fungus, it is easy to segment out background artifacts manually. Each generated slice chooses a background from the set of 30 different manually segmented backgrounds. Finally, zero-mean Gaussian noise with $0.001$ variance is added. 

\begin{table}[!ht]
\centering
\small
\caption{Comparison of different algorithms for segmentation on synthetic data. Precision, recall, and F1 are average percentages, while time is measured in average seconds to process a single 500 x 500 x 150 image stack.}
\label{tab:seg_compare}
\begin{tabular}{|c|c|c|c|c|}
\hline
Algorithm  & Prec & Recall  & F1 & Time(s) \\ \hline
Deep CNN \cite{DBLP:journals/corr/MajiSMS16} & 21.23 & 97.5 & \textbf{34.87} & 2700 \\ 
Frangi \cite{frangi:2001-53} & \textbf{26.08} & 51.31 & 34.58 & 109.48 \\
Var3d \cite{doi:10.1093/bioinformatics/btt276} & 6.04 & \textbf{99.86} & 11.39 & 6.83 \\
Phansalkar \cite{5739305} & 0.05 & 100 & 0.09 & 8.25 \\
Frang+Phan & 6.14 & 95.08& 11.54& 124.04 \\
Coye \cite{coye} & 2.85 & 14.2788 & 4.7583 & \textbf{3.6574} \\
\hline
\end{tabular}
\vspace{-1em}
\end{table}

\subsubsection{Quantitative Analysis with Synthetically Generated Image}
For these experiments, 100 fungal networks were synthetically generated containing on average $1,054$ voxels each. Generated image stacks contained a random number between [1,5] of fungal networks and are 500 x 500 x 150 in resolution. All algorithms were run on a Sager Laptop containing an Intel Core i7-6400k CPU and 32GB of RAM, and a Nvidia 1080 GPU.

The metrics we used for comparison are precision, recall, and F1 score. $Prec = \frac{tp}{tp+fp}$, which measures the probability that segmented voxels correspond to hyphae ($tp$ or true positive) versus not hyphae ($fp$ or true false positive). $Recall = \frac{tp}{tp+fn}$, which measures the probability that true hyphal voxels are segmented, given that each of the $tp$ and $fp$ voxels can be defined a prior. In other words, precision measures how well a segmentation scheme extracts only fungus, whereas recall measures the amount of fungus we actually segment. The F1 score is a measurement that combines both precision and recall via harmonic mean. That is $F1 = 2*\frac{prec*recall}{prec+recall}$. The F1 metric is a good way to use one number to compare segmentations, whereas the precision and recall individually show sources of the error. We choose precision/recall as a statistic because they are more representative when the amount of positives are much fewer than the amount of negatives. For each image stack there are $37,500,000$ voxels, but only up to 5270 contain fungus.
%
%

%

We tested a variety of algorithms. We attempted to use many filters and segmentation schemes available on ImageJ \cite{schindelin2015imagej, schindelin2012fiji}, an open platform for biological or medical image processing. Since some methods will output a binary image stack and some methods output a grayscale confidence, we threshold such that the F1 score is maximized.

As can be seen in Table \ref{tab:seg_compare}, the Frangi filter with an optimal threshold and the deep CNN have comparable F1 scores. However, the deep CNN directly outputs a segmentation, while the Frangi filtered must be thresholded, and finding the optimal threshold automatically adds another challenge. The downside of the deep CNN is that it can take 45 minutes to process an entire stack even when using a recent GPU. We tried a few auto-thresholding techniques, and the Phansalkar method produced the best results. However, it still completely fails due to noise and background artifacts, even after filtering the image stack first. 

\vspace{-1em}
\subsubsection{Quantitative Analysis of Gap Closing Algorithm}
\begin{figure}[!htbp]
\centering
\includegraphics[width=0.45\textwidth]{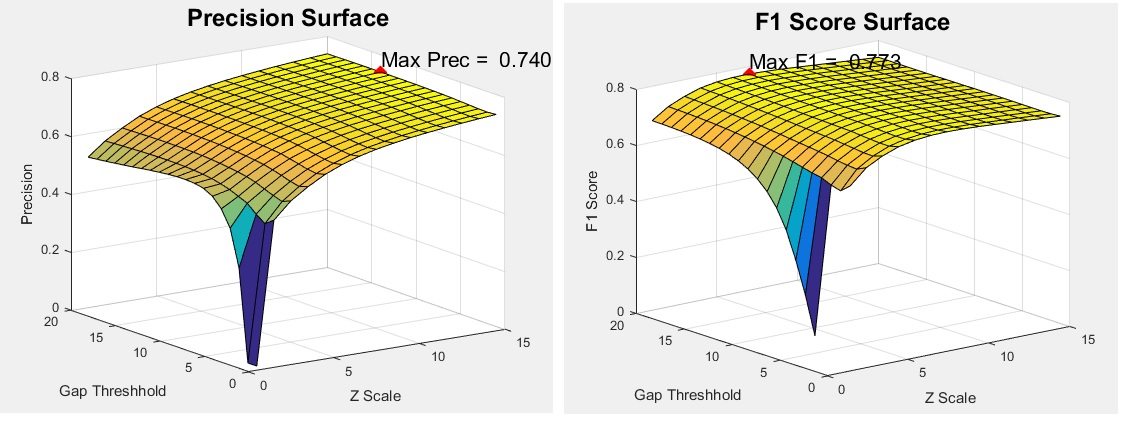}
\caption{A precision and F1 score surface created by varying the gap length and z-scale parameters for the skeleton gap algorithm. The peak on the precision surface is 74\% while the peak on the F1 surface is 77.3\%}
\label{prec_surface}
\end{figure}

\vspace{-.5em}
To test the gap closing algorithm, we take the ground truth generated hyphal skeleton, and for each branch we add a random amount of gaps of size [1,10]. This is performed on 100 skeletons. We measure the effectiveness of our gap closing algorithm by comparing endpoints. A true positive corresponds to endpoints that were connected by the algorithm that were connected in the ground truth data, a true negative corresponds to endpoints that were left unconnected by the algorithm that were unconnected in the ground truth data, a false positive corresponds to endpoints that were connected by the algorithm that were not connected in the ground truth data, and a false negative corresponds to endpoints that were left unconnected by the algorithm that were connected in the ground truth data. A precision, recall, and F1 surface were calculated by taking an average over the 100 skeletons, and varying the gap length and z-scale parameters, shown in Figure \ref{prec_surface}. 

\vspace{-1em}
These surfaces show that the precision increases with z-scale, which implies most connections are in-plane for each slice. Recall increases with a higher gap length. F1 score is maximized with a z-scaling of 7 and maximal gap length.

\vspace{-1em}
\subsection{Qualitative Analysis with Real Data}
\vspace{-1em}
\label{sec:real}
We use images described by Minker et al. \cite{minker2016semiautomated}. Briefly, 100 image stacks per leaf punch were taken in a 10x10 grid using Laser scanning confocal microscopy on a Zeiss LSM5 DUO. Samples were cleared and stained to highlight the fungal hyphae. Shade correction was performed to remove intensity variation between each image stack, and stitching was performed to create a full leaf reconstruction at microscopic detail -- i.e. a macroscopic microscopy image stack. We show the results on 3 slices with a small, medium, and large amount of fungus, respectively. We have run these algorithms on more data and have observed that the deep CNN perform the best, although some image stacks have strong artifacts which cause it to fail. 

\begin{figure}[!htbp]
\centering
\includegraphics[width=0.40\textwidth]{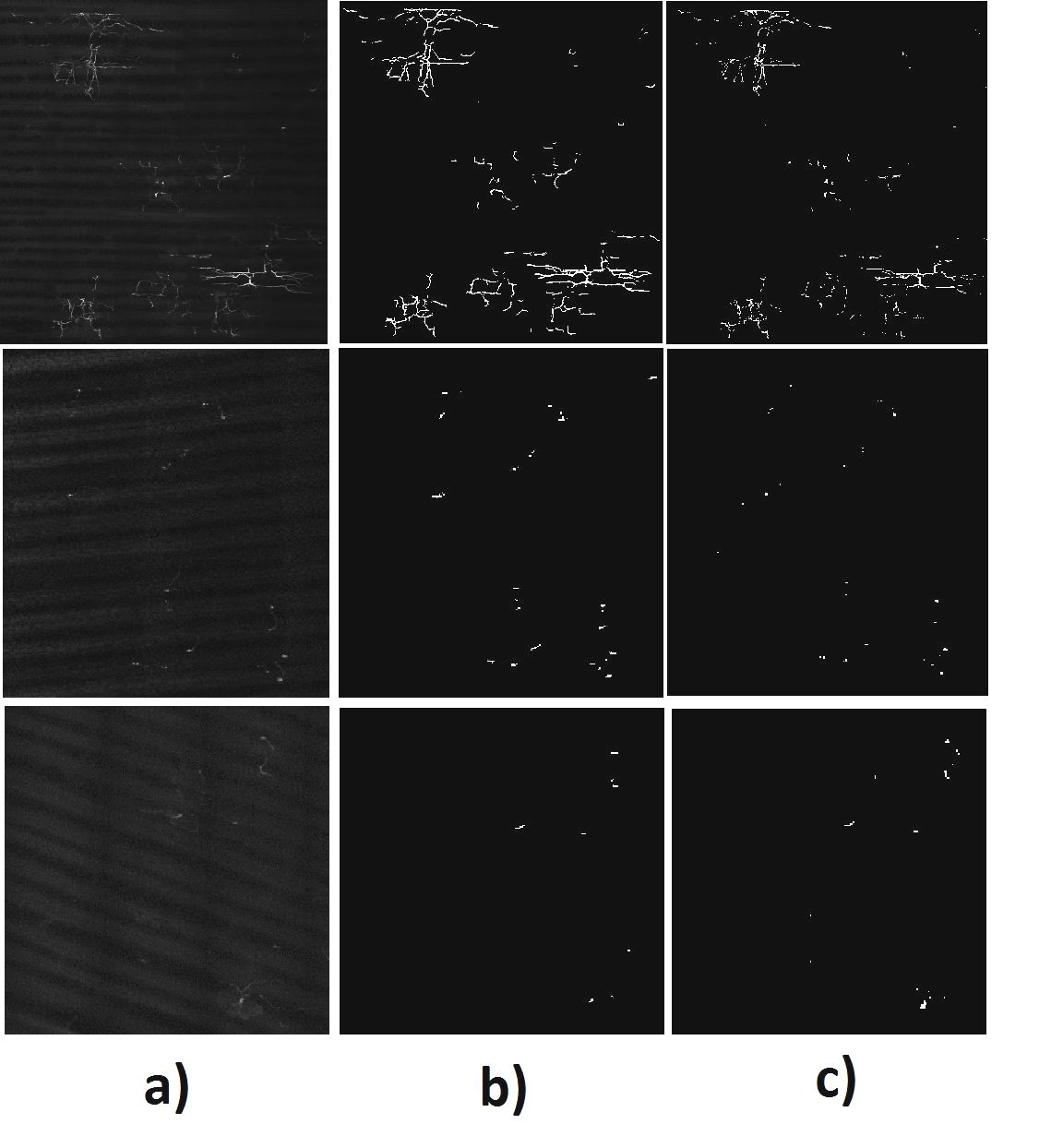}
\vspace{-1em}
\caption{Comparison of single image slices from 3 different real microscopy image stacks. \textbf{a)} Raw image slice \textbf{b)}  Segmentation with the deep learning method described in \cite{DBLP:journals/corr/MajiSMS16} \textbf{b)}  Segmentation with the frangi filter and threshold of 0.05 \cite{frangi:2001-53}}
\label{real_qual}

\end{figure}

\vspace{-2em}
\section{Conclusion}
\vspace{-1em}
We created a synthetic fungal hyphal generator which creates real 3D image stacks that mimics the properties of real data. We compared segmentation and filtering methods on our data. The deep CNN method of \cite{DBLP:journals/corr/MajiSMS16} produced the best results, but also took the most amount of time to process. We also compare the deep CNN method against the frangi filter on real microscopy data, and found that the deep CNN produced good results overall. Real data can contain artifacts which cause small gaps to appear in even the best segmentation methods. To handle this, we developed a skeleton gap closing algorithm based on a minimum spanning tree. We used the synthetic fungal generator to quantitative analyze the performance of the gap closing algorithm with synthetic gaps and achieved a maximal F1 score of 77.3 \%. In the future, we will develop new segmentation algorithms to run on the real microscopy data for automatic fungal resistance analysis.

\label{sec:con}

\bibliographystyle{IEEEbib}
\bibliography{strings,refs}

\end{document}